\PassOptionsToPackage{unicode}{hyperref}
\PassOptionsToPackage{hyphens}{url}
\documentclass[
]{article}
\usepackage{amsmath,amssymb}
\usepackage{iftex}
\ifPDFTeX
  \usepackage[T1]{fontenc}
  \usepackage[utf8]{inputenc}
  \usepackage{textcomp} 
\else 
  \usepackage{unicode-math} 
  \defaultfontfeatures{Scale=MatchLowercase}
  \defaultfontfeatures[\rmfamily]{Ligatures=TeX,Scale=1}
\fi
\usepackage{lmodern}
\ifPDFTeX\else
\fi
\IfFileExists{upquote.sty}{\usepackage{upquote}}{}
\IfFileExists{microtype.sty}{
  \usepackage[]{microtype}
  \UseMicrotypeSet[protrusion]{basicmath} 
}{}
\makeatletter
\@ifundefined{KOMAClassName}{
  \IfFileExists{parskip.sty}{%
    \usepackage{parskip}
  }{
    \setlength{\parindent}{0pt}
    \setlength{\parskip}{6pt plus 2pt minus 1pt}}
}{
  \KOMAoptions{parskip=half}}
\makeatother
\usepackage{xcolor}
\usepackage[margin=1in]{geometry}
\usepackage{color}
\usepackage{fancyvrb}

\DefineVerbatimEnvironment{Highlighting}{Verbatim}{commandchars=\\\{\}}
\usepackage{framed}
\definecolor{shadecolor}{RGB}{248,248,248}
\newenvironment{Shaded}{\begin{snugshade}}{\end{snugshade}}

\newcommand{\AttributeTok}[1]{\textcolor[rgb]{0.13,0.29,0.53}{#1}}

\newcommand{\BuiltInTok}[1]{#1}

\newcommand{\CommentTok}[1]{\textcolor[rgb]{0.56,0.35,0.01}{\textit{#1}}}

\newcommand{\ControlFlowTok}[1]{\textcolor[rgb]{0.13,0.29,0.53}{\textbf{#1}}}

\newcommand{\DecValTok}[1]{\textcolor[rgb]{0.00,0.00,0.81}{#1}}

\newcommand{\FloatTok}[1]{\textcolor[rgb]{0.00,0.00,0.81}{#1}}
\newcommand{\FunctionTok}[1]{\textcolor[rgb]{0.13,0.29,0.53}{\textbf{#1}}}
\newcommand{\ImportTok}[1]{#1}

\newcommand{\KeywordTok}[1]{\textcolor[rgb]{0.13,0.29,0.53}{\textbf{#1}}}
\newcommand{\NormalTok}[1]{#1}
\newcommand{\OperatorTok}[1]{\textcolor[rgb]{0.81,0.36,0.00}{\textbf{#1}}}
\newcommand{\OtherTok}[1]{\textcolor[rgb]{0.56,0.35,0.01}{#1}}

\newcommand{\SpecialCharTok}[1]{\textcolor[rgb]{0.81,0.36,0.00}{\textbf{#1}}}

\newcommand{\StringTok}[1]{\textcolor[rgb]{0.31,0.60,0.02}{#1}}
\newcommand{\VariableTok}[1]{\textcolor[rgb]{0.00,0.00,0.00}{#1}}

\usepackage{longtable,booktabs,array}
\usepackage{calc} 
\usepackage{etoolbox}
\makeatletter
\patchcmd\longtable{\par}{\if@noskipsec\mbox{}\fi\par}{}{}
\makeatother
\IfFileExists{footnotehyper.sty}{\usepackage{footnotehyper}}{\usepackage{footnote}}
\makesavenoteenv{longtable}
\usepackage{graphicx}
\makeatletter
\def\maxwidth{\ifdim\Gin@nat@width>\linewidth\linewidth\else\Gin@nat@width\fi}
\def\maxheight{\ifdim\Gin@nat@height>\textheight\textheight\else\Gin@nat@height\fi}
\makeatother
\setkeys{Gin}{width=\maxwidth,height=\maxheight,keepaspectratio}
\makeatletter
\def\fps@figure{htbp}
\makeatother
\providecommand{\tightlist}{%
  \setlength{\itemsep}{0pt}\setlength{\parskip}{0pt}}
\newlength{\cslhangindent}
\newlength{\csllabelwidth}
\newlength{\cslentryspacingunit} 
\newenvironment{CSLReferences}[2] 
 {
  \setlength{\parindent}{0pt}
  \ifodd #1
  \let\oldpar\par
  \def\par{\hangindent=\cslhangindent\oldpar}
  \fi
  \setlength{\parskip}{#2\cslentryspacingunit}
 }%
 {}
\usepackage{calc}

\ifLuaTeX
  \usepackage{selnolig}  
\fi
\IfFileExists{bookmark.sty}{\usepackage{bookmark}}{\usepackage{hyperref}}
\IfFileExists{xurl.sty}{\usepackage{xurl}}{} 
\hypersetup{
  pdftitle={Deep Learning: A Tutorial},
  pdfauthor={Nick Polson; Vadim Sokolov},
  hidelinks,
  pdfcreator={LaTeX via pandoc}}

\title{Deep Learning: A Tutorial}
\author{Nick Polson\footnote{University of Chicago,
  \href{mailto:ngp@chicagobooth.edu}{\nolinkurl{ngp@chicagobooth.edu}}} \and Vadim
Sokolov\footnote{George Mason University,
  \href{mailto:vsokolov@gmu.edu}{\nolinkurl{vsokolov@gmu.edu}}}}
\date{2023-10-09}

\begin{document}
\maketitle

\hypertarget{introduction}{%
\section{Introduction}\label{introduction}}

Our goal is to provide a review of deep learning methods which provide
insight into structured high-dimensional data. Rather than using shallow
additive architectures common to most statistical models, deep learning
uses layers of semi-affine input transformations to provide a predictive
rule. Applying these layers of transformations leads to a set of
attributes (or, features) to which probabilistic statistical methods can
be applied. Thus, the best of both worlds can be achieved: scalable
prediction rules fortified with uncertainty quantification, where sparse
regularization finds the features.

Deep learning is one of the widely used machine learning method for
analysis of large scale and high-dimensional data sets. Large-scale
means that we have many samples (observations) and high dimensional
means that each sample is a vector with many entries, usually hundreds
and up.

Machine learning is the engineer's version of statistical data analysis.
Major difference between ML and statistics is that ML focuses on
practical aspects, such as computational efficiency and ease of use of
techniques. While statistical analysis is more concerned with
rigorousness of the analysis and interpretability of the results.

Deep learning provides a powerful pattern matching tool suitable for
many AI applications. Image recognition and text analysis are probably
two of the deep learning's most successful. From a computational
perspective, you can think of an image or a text as a high dimensional
matrices and vectors, respectively. The problem of recognizing objects
in images or translating a text requires designing complex decision
boundaries in the high dimensional space of inputs.

Although, image analysis and natural language processing are the
applications where deep learning is the dominating approach, more
traditional engineering and science applications, such as
spatio-temporal and financial analysis is where DL also showed superior
performance compared to traditional statistical learning techniques
(Heaton et al. 2017; Polson and Sokolov 2017, 2023; Sokolov 2017; Dixon
et al. 2019; Polson and Sokolov 2020; Behnia et al. 2021; Bhadra et al.
2021; Polson et al. 2021; Nareklishvili et al. 2022a, b; 2023; Wang et
al. 2022)

There are several deep learning architectures exist - each has its own
uses and purposes. Convolutional Neural Networks (CNN) deal with
2-dimensional input objects, i.e.~images and were shown to outperform
any other techniques. Recurrent Neural Networks (RNN) were shown the
best performance on speech and text analysis tasks.

In general, a neural network can be described as follows. Let
\(f_1 , \ldots , f_L\) be given \emph{univariate} activation functions
for each of the \(L\) layers. Activation functions are nonlinear
transformations of weighted data. A semi-affine activation rule is then
defined by \[
f_l^{W,b} = f_l \left ( \sum_{j=1}^{N_l} W_{lj} X_j + b_l \right ) = f_l ( W_l X_l + b_l )\,,
\] which implicitly needs the specification of the number of hidden
units \(N_l\). Our deep predictor, given the number of layers \(L\),
then becomes the composite map

\[
\hat{Y}(X) = F(X) = \left ( f_l^{W_1,b_1} \circ \ldots \circ f_L^{W_L,b_L} \right ) ( X)\,.
\]

The fact that DL forms a universal `basis' which we recognise in this
formulation dates to Poincare and Hilbert is central. From a practical
perspective, given a large enough data set of ``test cases'', we can
empirically learn an optimal predictor.

Similar to a classic basis decomposition, the deep approach uses
univariate activation functions to decompose a high dimensional \(X\).

Let \(Z^{(l)}\) denote the \(l\)th layer, and so \(X = Z^{(0)}\). The
final output \(Y\) can be numeric or categorical. The explicit structure
of a deep prediction rule is then \[
\begin{aligned}
\hat{Y} (X) & = W^{(L)} Z^{(L)} + b^{(L)} \\
Z^{(1)} & = f^{(1)} \left ( W^{(0)} X + b^{(0)} \right ) \\
Z^{(2)} & = f^{(2)} \left ( W^{(1)} Z^{(1)} + b^{(1)} \right ) \\
\ldots  & \\
Z^{(L)} & = f^{(L)} \left ( W^{(L-1)} Z^{(L-1)} + b^{(L-1)} \right )\,.
\end{aligned}
\] Here \(W^{(l)}\) is a weight matrix and \(b^{(l)}\) are threshold or
activation levels. Designing a good predictor depends crucially on the
choice of univariate activation functions \(f^{(l)}\). The \(Z^{(l)}\)
are hidden features which the algorithm will extract.

Put differently, the deep approach employs hierarchical predictors
comprising of a series of \(L\) nonlinear transformations applied to
\(X\). Each of the \(L\) transformations is referred to as a
\emph{layer}, where the original input is \(X\), the output of the first
transformation is the first layer, and so on, with the output \(\hat Y\)
as the first layer. The layers \(1\) to \(L\) are called
\emph{hidden layers}. The number of layers \(L\) represents the
\emph{depth} of our routine.

Traditional statistical models are estimated by maximizing likelihood
and using the least squares algorithm for linear regression and weighted
least squares or Broyden-Fletcher-Goldfarb-Shanno (BFGS) algorithm for
generalized linear models.

\hypertarget{deep-learning-and-least-squares}{%
\section{Deep Learning and Least
Squares}\label{deep-learning-and-least-squares}}

The deep learning model approximates the relation between inputs \(x\)
and outputs \(y\) using a non-linear function \(f(x,\theta)\), where
\(\theta\) is a vector of parameters. The goal is to find the optimal
value of \(\theta\) that minimizes the expected loss function, given a
training data set \(D = \{x_i,y_i\}_{i=1}^n\). The loss function is a
measure of discrepancy between the true value of \(y\) and the predicted
value \(f(x,\theta)\). The loss function is usually defined as the
negative log-likelihood function of the model. \[
    l(\theta) = - \sum_{i=1}^n \log p(y_i | x_i, \theta),
\] where \(p(y_i | x_i, \theta)\) is the conditional distribution of
\(y_i\) given \(x_i\) and \(\theta\). Thus, in the case of regression,
we have \[
  y_i = f(x_i,\theta) + \epsilon, ~ \epsilon \sim N(0,\sigma^2),
\] Thus, the loss function is \[
    l(\theta) = - \sum_{i=1}^n \log p(y_i | x_i, \theta) = \sum_{i=1}^n (y_i - f(x_i, \theta))^2,
\]

\hypertarget{regresson}{%
\subsection{Regresson}\label{regresson}}

Regression is simply a neural network which is wide and shallow. The
insight of DL is that you use a deep and shallow neural network. Let's
look at a simple example and fit a linear regression model to
\texttt{iris} dataset.

\begin{Shaded}
\begin{Highlighting}[]
\FunctionTok{data}\NormalTok{(iris)}
\NormalTok{y }\OtherTok{=}\NormalTok{ iris}\SpecialCharTok{$}\NormalTok{Petal.Length}
\CommentTok{\# initialize theta}
\NormalTok{theta }\OtherTok{\textless{}{-}} \FunctionTok{matrix}\NormalTok{(}\FunctionTok{c}\NormalTok{(}\DecValTok{0}\NormalTok{, }\DecValTok{0}\NormalTok{), }\AttributeTok{nrow =} \DecValTok{2}\NormalTok{, }\AttributeTok{ncol =} \DecValTok{1}\NormalTok{)}
\CommentTok{\# learning rate}
\NormalTok{alpha }\OtherTok{\textless{}{-}} \FloatTok{0.0001}
\CommentTok{\# number of iterations}
\NormalTok{n\_iter }\OtherTok{\textless{}{-}} \DecValTok{1000}
\CommentTok{\# gradient descent}
\ControlFlowTok{for}\NormalTok{ (i }\ControlFlowTok{in} \DecValTok{1}\SpecialCharTok{:}\NormalTok{n\_iter) \{}
    \CommentTok{\# compute gradient}
\NormalTok{    grad }\OtherTok{\textless{}{-}} \SpecialCharTok{{-}}\DecValTok{2} \SpecialCharTok{*} \FunctionTok{t}\NormalTok{(x) }\SpecialCharTok{\%*\%}\NormalTok{ (y }\SpecialCharTok{{-}}\NormalTok{ x }\SpecialCharTok{\%*\%}\NormalTok{ theta)}
    \CommentTok{\# update theta}
\NormalTok{    theta }\OtherTok{\textless{}{-}}\NormalTok{ theta }\SpecialCharTok{{-}}\NormalTok{ alpha }\SpecialCharTok{*}\NormalTok{ grad}
\NormalTok{\}}
\end{Highlighting}
\end{Shaded}

Our gradient descent finds the following coefficients

\begin{longtable}[]{@{}rr@{}}
\toprule\noalign{}
Intercept (\(\theta_1\)) & Petal.Width (\(\theta_2\)) \\
\midrule\noalign{}
\endhead
\bottomrule\noalign{}
\endlastfoot
1.1 & 2.2 \\
\end{longtable}

Let's plot the data and model estimated using gradient descent

\begin{Shaded}
\begin{Highlighting}[]
\FunctionTok{plot}\NormalTok{(x[,}\DecValTok{2}\NormalTok{],y,}\AttributeTok{pch=}\DecValTok{16}\NormalTok{, }\AttributeTok{xlab=}\StringTok{"Petal.Width"}\NormalTok{)}
\FunctionTok{abline}\NormalTok{(theta[}\DecValTok{2}\NormalTok{],theta[}\DecValTok{1}\NormalTok{], }\AttributeTok{lwd=}\DecValTok{3}\NormalTok{,}\AttributeTok{col=}\StringTok{"red"}\NormalTok{)}
\end{Highlighting}
\end{Shaded}

\begin{center}\includegraphics[width=0.7\linewidth]{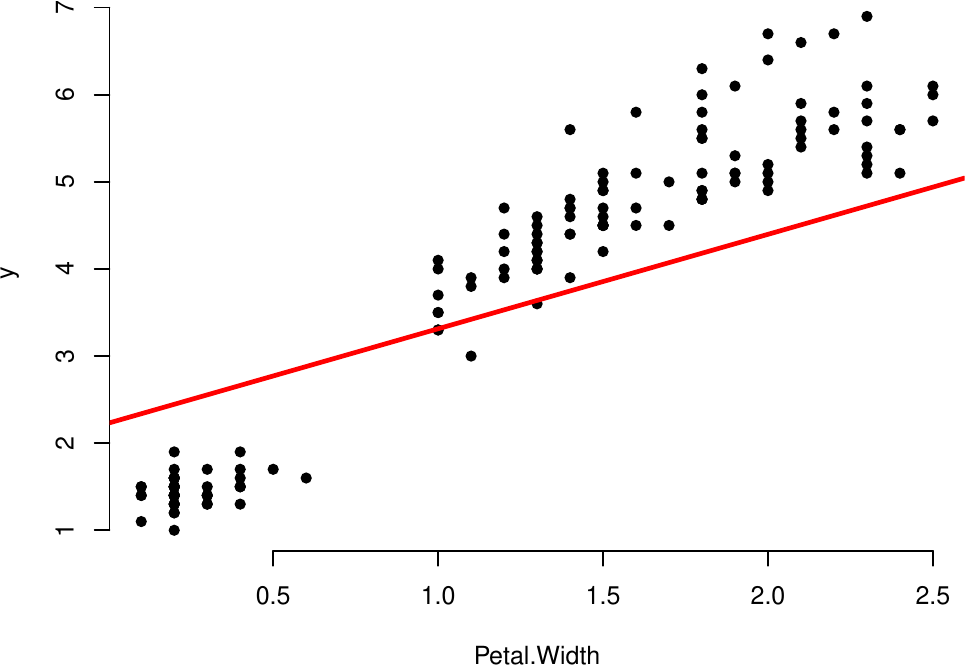} \end{center}

Let's compare it to the standard estimation algorithm

\begin{Shaded}
\begin{Highlighting}[]
\NormalTok{m }\OtherTok{=} \FunctionTok{lm}\NormalTok{(Petal.Length}\SpecialCharTok{\textasciitilde{}}\NormalTok{Petal.Width, }\AttributeTok{data=}\NormalTok{iris)}
\end{Highlighting}
\end{Shaded}

\begin{longtable}[]{@{}rr@{}}
\toprule\noalign{}
(Intercept) & Petal.Width \\
\midrule\noalign{}
\endhead
\bottomrule\noalign{}
\endlastfoot
1.1 & 2.2 \\
\end{longtable}

The values found by gradient descent are very close to the ones found by
the standard OLS algorithm.

\hypertarget{logistic-regression}{%
\subsection{Logistic Regression}\label{logistic-regression}}

Logistic regression is a generalized linear model (GLM) with a logit
link function, defined as: \[
    \log \left(\frac{p}{1-p}\right) = \theta_0 + \theta_1 x_1 + \ldots + \theta_p x_p,
\] where \(p\) is the probability of the positive class. The negative
log-likelihood function for logistic regression is a cross-entropy loss
\[
    l(\theta) = - \sum_{i=1}^n \left[ y_i \log p_i + (1-y_i) \log (1-p_i) \right],
\] where
\(p_i = 1/\left(1 + \exp(-\theta_0 - \theta_1 x_{i1} - \ldots - \theta_p x_{ip})\right)\).
The derivative of the negative log-likelihood function is \[
    \nabla l(\theta) = - \sum_{i=1}^n \left[ y_i - p_i \right] \begin{pmatrix} 1 \\ x_{i1} \\ \vdots \\ x_{ip} \end{pmatrix}.
\] In matrix notations, we have \[
    \nabla l(\theta) = - X^T (y - p).
\] Let's implement gradient descent algorithm now.

\begin{Shaded}
\begin{Highlighting}[]
\NormalTok{y }\OtherTok{=} \FunctionTok{ifelse}\NormalTok{(iris}\SpecialCharTok{$}\NormalTok{Species}\SpecialCharTok{==}\StringTok{"setosa"}\NormalTok{,}\DecValTok{1}\NormalTok{,}\DecValTok{0}\NormalTok{)}
\NormalTok{x }\OtherTok{=} \FunctionTok{cbind}\NormalTok{(}\FunctionTok{rep}\NormalTok{(}\DecValTok{1}\NormalTok{,}\DecValTok{150}\NormalTok{),iris}\SpecialCharTok{$}\NormalTok{Sepal.Length)}
\NormalTok{lrgd  }\OtherTok{=} \ControlFlowTok{function}\NormalTok{(x,y, alpha, n\_iter) \{}
\NormalTok{  theta }\OtherTok{\textless{}{-}} \FunctionTok{matrix}\NormalTok{(}\FunctionTok{c}\NormalTok{(}\DecValTok{0}\NormalTok{, }\DecValTok{0}\NormalTok{), }\AttributeTok{nrow =} \DecValTok{2}\NormalTok{, }\AttributeTok{ncol =} \DecValTok{1}\NormalTok{)}
  \ControlFlowTok{for}\NormalTok{ (i }\ControlFlowTok{in} \DecValTok{1}\SpecialCharTok{:}\NormalTok{n\_iter) \{}
    \CommentTok{\# compute gradient}
\NormalTok{    p }\OtherTok{=} \DecValTok{1}\SpecialCharTok{/}\NormalTok{(}\DecValTok{1}\SpecialCharTok{+}\FunctionTok{exp}\NormalTok{(}\SpecialCharTok{{-}}\NormalTok{x }\SpecialCharTok{\%*\%}\NormalTok{ theta))}
\NormalTok{    grad }\OtherTok{\textless{}{-}} \SpecialCharTok{{-}}\FunctionTok{t}\NormalTok{(x) }\SpecialCharTok{\%*\%}\NormalTok{ (y }\SpecialCharTok{{-}}\NormalTok{ p)}
    \CommentTok{\# update theta}
\NormalTok{    theta }\OtherTok{\textless{}{-}}\NormalTok{ theta }\SpecialCharTok{{-}}\NormalTok{ alpha }\SpecialCharTok{*}\NormalTok{ grad}
\NormalTok{  \}}
  \FunctionTok{return}\NormalTok{(theta)}
\NormalTok{\}}
\NormalTok{theta }\OtherTok{=} \FunctionTok{lrgd}\NormalTok{(x,y,}\FloatTok{0.005}\NormalTok{,}\DecValTok{20000}\NormalTok{)}
\end{Highlighting}
\end{Shaded}

The gradient descent parameters are

\begin{longtable}[]{@{}rr@{}}
\toprule\noalign{}
Intercept (\(\theta_1\)) & Sepal.Length (\(\theta_2\)) \\
\midrule\noalign{}
\endhead
\bottomrule\noalign{}
\endlastfoot
28 & -5.2 \\
\end{longtable}

And the plot is

\begin{Shaded}
\begin{Highlighting}[]
\FunctionTok{par}\NormalTok{(}\AttributeTok{mar=}\FunctionTok{c}\NormalTok{(}\DecValTok{4}\NormalTok{,}\DecValTok{4}\NormalTok{,}\DecValTok{0}\NormalTok{,}\DecValTok{0}\NormalTok{), }\AttributeTok{bty=}\StringTok{\textquotesingle{}n\textquotesingle{}}\NormalTok{)}
\FunctionTok{plot}\NormalTok{(x[,}\DecValTok{2}\NormalTok{],y,}\AttributeTok{pch=}\DecValTok{16}\NormalTok{, }\AttributeTok{xlab=}\StringTok{"Sepal.Length"}\NormalTok{)}
\FunctionTok{lines}\NormalTok{(x[,}\DecValTok{2}\NormalTok{],p,}\AttributeTok{type=}\StringTok{\textquotesingle{}p\textquotesingle{}}\NormalTok{, }\AttributeTok{pch=}\DecValTok{16}\NormalTok{,}\AttributeTok{col=}\StringTok{"red"}\NormalTok{)}
\end{Highlighting}
\end{Shaded}

\begin{center}\includegraphics[width=0.7\linewidth]{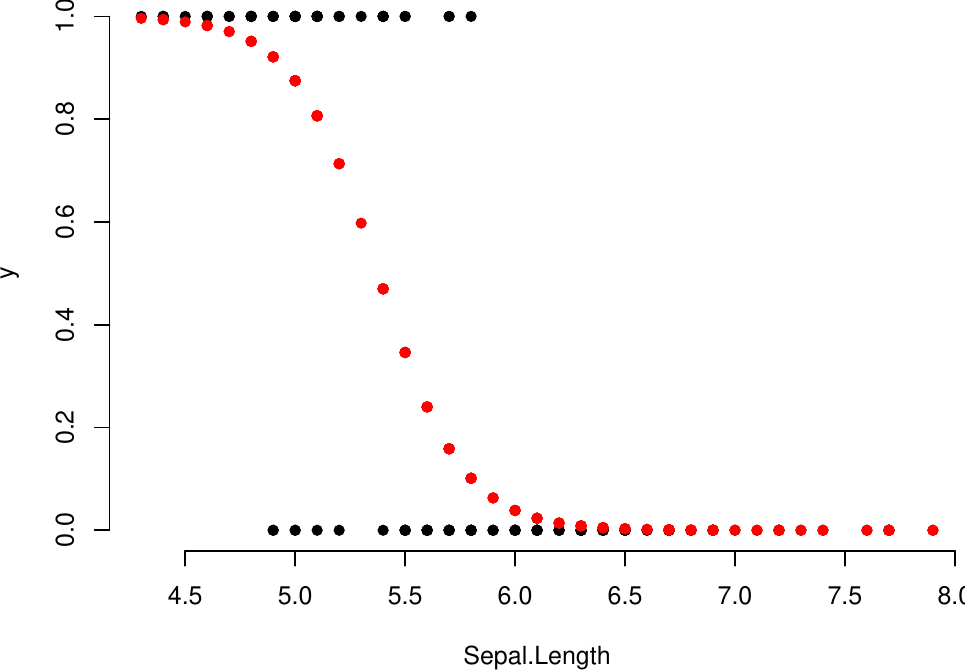} \end{center}

Let's compare it to the standard estimation algorithm

\begin{Shaded}
\begin{Highlighting}[]
\FunctionTok{glm}\NormalTok{(y}\SpecialCharTok{\textasciitilde{}}\NormalTok{x}\DecValTok{{-}1}\NormalTok{, }\AttributeTok{family=}\FunctionTok{binomial}\NormalTok{(}\AttributeTok{link=}\StringTok{"logit"}\NormalTok{))}
\CommentTok{\#\textgreater{} }
\CommentTok{\#\textgreater{} Call:  glm(formula = y \textasciitilde{} x {-} 1, family = binomial(link = "logit"))}
\CommentTok{\#\textgreater{} }
\CommentTok{\#\textgreater{} Coefficients:}
\CommentTok{\#\textgreater{}    x1     x2  }
\CommentTok{\#\textgreater{} 27.83  {-}5.18  }
\CommentTok{\#\textgreater{} }
\CommentTok{\#\textgreater{} Degrees of Freedom: 150 Total (i.e. Null);  148 Residual}
\CommentTok{\#\textgreater{} Null Deviance:       208 }
\CommentTok{\#\textgreater{} Residual Deviance: 72    AIC: 76}
\end{Highlighting}
\end{Shaded}

\hypertarget{model-estimation}{%
\subsection{Model Estimation}\label{model-estimation}}

Deep learning as well as a large number of statistical problems, can be
expressed in the form \[
\min l(x) + \phi(x).
\] In learning \(l(x)\) is the negative log-likelihood and \(\phi(x)\)
is a penalty function that regularizes the estimate. From the Bayesian
perspective, the solution to this problem may be interpreted as a
maximum a posteriori
\[p(y\mid x) \propto \exp\{-l(x)\}, ~ p(x) \propto \exp\{-\phi(x)\}.\]

Second order optimisation algorithms, such as BFGS used for traditional
statistical models do not work well for deep learning models. The reason
is that the number of parameters a DL model has is large and estimating
second order derivatives (Hessian or Fisher information matrix) becomes
prohibitive from both computational and memory use standpoints. Instead,
first order gradient descent methods are used for estimating parameters
of a deep learning models.

The problem of parameter estimation (when likelihood belongs to the
exponential family) is an optimisation problem

\[
    \min_{\theta} l(\theta) := \dfrac{1}{n} \sum_{i=1}^n \log p(y_i, f(x_i, \theta))
\] where \(l\) is the negative log-likelihood of a sample, and
\(\theta\) is the vector of parameters. The gradient descent method is
an iterative algorithm that starts with an initial guess \(\theta^{0}\)
and then updates the parameter vector \(\theta\) at each iteration \(t\)
as follows: \[
    \theta^{t+1} = \theta^t - \alpha_t \nabla l(\theta^t).
\]

Let's demonstrate these algorithms on a simple example of linear
regression. We will use the \texttt{mtcars} data set and try to predict
the fuel consumption (mpg) \(y\) using the number of cylinders (cyl) as
a predictor \(x\). We will use the following model: \[
    y_i = \theta_0 + \theta_1 x_i + \epsilon_i,
\] or in matrix form \[
    y = X \theta + \epsilon,
\] where \(\epsilon_i \sim N(0, \sigma^2)\), \(X = [1 ~ x]\) is the
design matrix with first column beign all ones.

The negative log-likelihood function for the linear regression model is
\[
    l(\theta) = \sum_{i=1}^n (y_i - \theta_0 - \theta_1 x_i)^2.
\]

The gradient then is \[
    \nabla l(\theta) = -2 \sum_{i=1}^n (y_i - \theta_0 - \theta_1 x_i) \begin{pmatrix} 1 \\ x_i \end{pmatrix}.
\] In matrix form, we have \[
    \nabla l(\theta) = -2 X^T (y - X \theta).   
\]

Now, we demonstrate the gradient descent for estimating a generalized
linear model (GLM), namely logistic regression. We will use the
\texttt{iris} data set again and try to predict the species of the
flower using the petal width as a predictor. We will use the following
model \[
    \log \left(\frac{p_i}{1-p_i}\right) = \theta_0 + \theta_1 x_i,
\] where \(p_i = P(y_i = 1)\) is the probability of the flower being of
the species \(y_i = 1\) (setosa).

The negative log-likelihood function for logistic regression model is \[
    l(\theta) = - \sum_{i=1}^n \left[ y_i \log p_i + (1-y_i) \log (1-p_i) \right],
\] where \(p_i = 1/\left(1 + \exp(-\theta_0 - \theta_1 x_i)\right)\).
The derivative of the negative log-likelihood function is \[
    \nabla l(\theta) = - \sum_{i=1}^n \left[ y_i - p_i \right] \begin{pmatrix} 1 \\ x_i \end{pmatrix}.
\] In matrix notations, we have \[
    \nabla l(\theta) = - X^T (y - p).
\]

\hypertarget{stochastic-gradient-descent}{%
\subsection{Stochastic Gradient
Descent}\label{stochastic-gradient-descent}}

Stochastic gradient descent (SGD) is a variant of the gradient descent
algorithm. The main difference is that instead of computing the gradient
over the whole data set, SGD computes the gradient over a randomly
selected subset of the data. This allows SGD to be applied to estimate
models when data set is too large to fit into memory, which is often the
case with the deep learning models. The SGD algorithm replaces the
gradient of the negative log-likelihood function with the gradient of
the negative log-likelihood function computed over a randomly selected
subset of the data \[
    \nabla l(\theta) \approx \dfrac{1}{|B|} \sum_{i \in B} \nabla l(y_i, f(x_i, \theta)),
\] where \(B \in \{1,2,\ldots,n\}\) is the batch samples from the data
set. This method can be interpreted as gradient descent using noisy
gradients, which are typically called mini-batch gradients with batch
size \(|B|\).

The SGD is based on the idea of stochastic approximation introduced by
Robbins and Monro (1951). Stochastic simply replaces \(F(l)\) with its
Monte Carlo approximation.

In a small mini-batch regime, when \(|B| \ll n\) and typically
\(|B| \in \{32,64,\ldots,1024\}\) it was shown that SGD converges faster
than the standard gradient descent algorithm, it does converge to
minimizers of strongly convex functions (negative log-likelihood
function from exponential family is strongly convex) (Bottou et al.
2018) and it is more robust to noise in the data (Hardt et al. 2016).
Further, it was shown that it can avoid saddle-points, which is often an
issue with deep learning log-likelihood functions. In the case of
multiple-minima, SGD can find a good solution LeCun et al. (2002),
meaning that the out-of-sample performance is often worse when trained
with large- batch methods as compared to small-batch methods.

Now, we implement SGD for logistic regression and compare performance
for different batch sizes

\begin{Shaded}
\begin{Highlighting}[]
\NormalTok{lrgd\_minibatch  }\OtherTok{=} \ControlFlowTok{function}\NormalTok{(x,y, alpha, n\_iter, bs) \{}
\NormalTok{  theta }\OtherTok{\textless{}{-}} \FunctionTok{matrix}\NormalTok{(}\FunctionTok{c}\NormalTok{(}\DecValTok{0}\NormalTok{, }\DecValTok{0}\NormalTok{), }\AttributeTok{nrow =} \DecValTok{2}\NormalTok{, }\AttributeTok{ncol =}\NormalTok{ n\_iter}\SpecialCharTok{+}\DecValTok{1}\NormalTok{)}
\NormalTok{  n }\OtherTok{=} \FunctionTok{length}\NormalTok{(y)}
  \ControlFlowTok{for}\NormalTok{ (i }\ControlFlowTok{in} \DecValTok{1}\SpecialCharTok{:}\NormalTok{n\_iter) \{}
\NormalTok{    s }\OtherTok{=}\NormalTok{ ((i}\DecValTok{{-}1}\NormalTok{)}\SpecialCharTok{*}\NormalTok{bs}\SpecialCharTok{+}\DecValTok{1}\NormalTok{)}\SpecialCharTok{\%\%}\NormalTok{n}
\NormalTok{    e }\OtherTok{=} \FunctionTok{min}\NormalTok{(s}\SpecialCharTok{+}\NormalTok{bs}\DecValTok{{-}1}\NormalTok{,n)}
\NormalTok{    xl }\OtherTok{=}\NormalTok{ x[s}\SpecialCharTok{:}\NormalTok{e,]; yl }\OtherTok{=}\NormalTok{ y[s}\SpecialCharTok{:}\NormalTok{e]}
\NormalTok{    p }\OtherTok{=} \DecValTok{1}\SpecialCharTok{/}\NormalTok{(}\DecValTok{1}\SpecialCharTok{+}\FunctionTok{exp}\NormalTok{(}\SpecialCharTok{{-}}\NormalTok{xl }\SpecialCharTok{\%*\%}\NormalTok{ theta[,i]))}
\NormalTok{    grad }\OtherTok{\textless{}{-}} \SpecialCharTok{{-}}\FunctionTok{t}\NormalTok{(xl) }\SpecialCharTok{\%*\%}\NormalTok{ (yl }\SpecialCharTok{{-}}\NormalTok{ p)}
    \CommentTok{\# update theta}
\NormalTok{    theta[,i}\SpecialCharTok{+}\DecValTok{1}\NormalTok{] }\OtherTok{\textless{}{-}}\NormalTok{ theta[,i] }\SpecialCharTok{{-}}\NormalTok{ alpha }\SpecialCharTok{*}\NormalTok{ grad}
\NormalTok{  \}}
  \FunctionTok{return}\NormalTok{(theta)}
\NormalTok{\}}
\end{Highlighting}
\end{Shaded}

Now run our SGD algorithm with different batch sizes.

\begin{Shaded}
\begin{Highlighting}[]
\FunctionTok{set.seed}\NormalTok{(}\DecValTok{92}\NormalTok{) }\CommentTok{\# kuzy}
\NormalTok{ind }\OtherTok{=} \FunctionTok{sample}\NormalTok{(}\DecValTok{150}\NormalTok{)}
\NormalTok{y }\OtherTok{=} \FunctionTok{ifelse}\NormalTok{(iris}\SpecialCharTok{$}\NormalTok{Species}\SpecialCharTok{==}\StringTok{"setosa"}\NormalTok{,}\DecValTok{1}\NormalTok{,}\DecValTok{0}\NormalTok{)[ind] }\CommentTok{\# shuffle data}
\NormalTok{x }\OtherTok{=} \FunctionTok{cbind}\NormalTok{(}\FunctionTok{rep}\NormalTok{(}\DecValTok{1}\NormalTok{,}\DecValTok{150}\NormalTok{),iris}\SpecialCharTok{$}\NormalTok{Sepal.Length)[ind,] }\CommentTok{\# shuffle data}
\NormalTok{nit}\OtherTok{=}\DecValTok{200000}
\NormalTok{lr }\OtherTok{=} \FloatTok{0.01}
\NormalTok{th1 }\OtherTok{=} \FunctionTok{lrgd\_minibatch}\NormalTok{(x,y,lr,nit,}\DecValTok{5}\NormalTok{)}
\NormalTok{th2 }\OtherTok{=} \FunctionTok{lrgd\_minibatch}\NormalTok{(x,y,lr,nit,}\DecValTok{15}\NormalTok{)}
\NormalTok{th3 }\OtherTok{=} \FunctionTok{lrgd\_minibatch}\NormalTok{(x,y,lr,nit,}\DecValTok{30}\NormalTok{)}
\end{Highlighting}
\end{Shaded}

\begin{center}\includegraphics[width=0.7\linewidth]{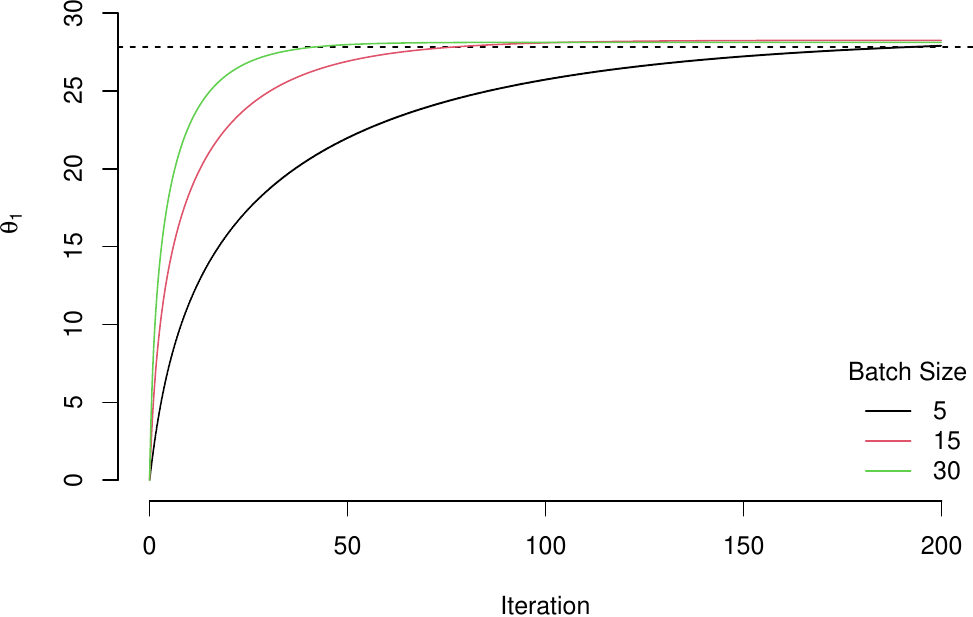} \end{center}

We run it with \ensuremath{2\times 10^{5}} iterations and the learning
rate of 0.01 and plot the values of \(\theta_1\) every 1000 iteration.
There are a couple of important points we need to highlight when using
SGD. First, we shuffle the data before using it. The reason is that if
the data is sorted in any way (e.g.~by date or by value of one of the
inputs), then data within batches can be highly correlated, which
reduces the convergence speed. Shuffling helps avoiding this issue.
Second, the larger the batch size, the smaller number of iterations are
required for convergence, which is something we would expect. However,
in this specific example, from the number of computation point of view,
the batch size does not change the number calculations required overall.
Let's look at the same plot, but scale the x-axis according to the
amount of computations

\begin{Shaded}
\begin{Highlighting}[]
\FunctionTok{plot}\NormalTok{(ind}\SpecialCharTok{/}\DecValTok{1000}\NormalTok{,th1[}\DecValTok{1}\NormalTok{,ind], }\AttributeTok{type=}\StringTok{\textquotesingle{}l\textquotesingle{}}\NormalTok{, }\AttributeTok{ylim=}\FunctionTok{c}\NormalTok{(}\DecValTok{0}\NormalTok{,}\DecValTok{33}\NormalTok{), }\AttributeTok{col=}\DecValTok{1}\NormalTok{, }\AttributeTok{ylab=}\FunctionTok{expression}\NormalTok{(theta[}\DecValTok{1}\NormalTok{]), }\AttributeTok{xlab=}\StringTok{"Iteration"}\NormalTok{)}
\FunctionTok{abline}\NormalTok{(}\AttributeTok{h=}\FloatTok{27.83}\NormalTok{, }\AttributeTok{lty=}\DecValTok{2}\NormalTok{)}
\FunctionTok{lines}\NormalTok{(ind}\SpecialCharTok{/}\DecValTok{1000}\SpecialCharTok{*}\DecValTok{3}\NormalTok{,th2[}\DecValTok{1}\NormalTok{,ind], }\AttributeTok{type=}\StringTok{\textquotesingle{}l\textquotesingle{}}\NormalTok{, }\AttributeTok{col=}\DecValTok{2}\NormalTok{)}
\FunctionTok{lines}\NormalTok{(ind}\SpecialCharTok{/}\DecValTok{1000}\SpecialCharTok{*}\DecValTok{6}\NormalTok{,th3[}\DecValTok{1}\NormalTok{,ind], }\AttributeTok{type=}\StringTok{\textquotesingle{}l\textquotesingle{}}\NormalTok{, }\AttributeTok{col=}\DecValTok{3}\NormalTok{)}
\FunctionTok{legend}\NormalTok{(}\StringTok{"bottomright"}\NormalTok{, }\AttributeTok{legend=}\FunctionTok{c}\NormalTok{(}\DecValTok{5}\NormalTok{,}\DecValTok{15}\NormalTok{,}\DecValTok{30}\NormalTok{),}\AttributeTok{col=}\DecValTok{1}\SpecialCharTok{:}\DecValTok{3}\NormalTok{, }\AttributeTok{lty=}\DecValTok{1}\NormalTok{, }\AttributeTok{bty=}\StringTok{\textquotesingle{}n\textquotesingle{}}\NormalTok{,}\AttributeTok{title =} \StringTok{"Batch Size"}\NormalTok{)}
\end{Highlighting}
\end{Shaded}

\begin{center}\includegraphics[width=0.7\linewidth]{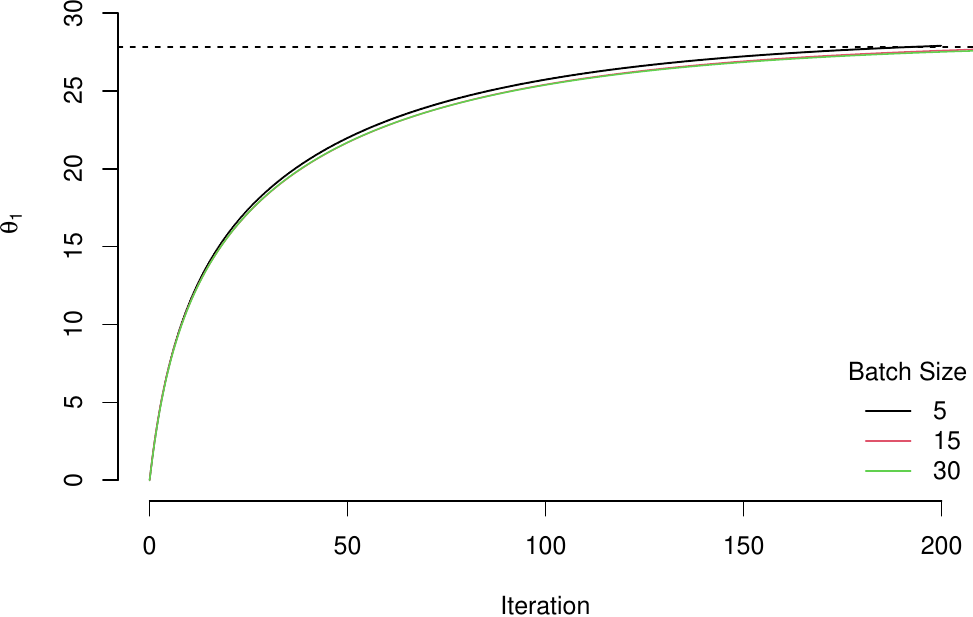} \end{center}

There are several important considerations about choosing the batch size
for SGD.

\begin{itemize}
\tightlist
\item
  The larger the batch size, the more memory is required to store the
  data.
\item
  Parallelization is more efficient with larger batch sizes. Modern
  harware supports parallelization of matrix operations, which is the
  main operation in SGD. The larger the batch size, the more efficient
  the parallelization is. Usually there is a sweet spot \(|B|\) for the
  batch size, which is the largest batch size that can fit into the
  memory or parallelized. Meaning it takes the same amount of time to
  compute SGD step for batch size \(1\) and \(B\).
\item
  Third, the larger the batch size, the less noise in the gradient. This
  means that the larger the batch size, the more accurate the gradient
  is. However, it was empirically shown that in many applications we
  should prefer noisier gradients (small batches) to obtain high quality
  solutions when the objective function (negative log-likelihood) is
  non-convex (Keskar et al. 2016).
\end{itemize}

\hypertarget{feed-forward-relu-neural-networks}{%
\section{Feed-Forward ReLu Neural
Networks}\label{feed-forward-relu-neural-networks}}

Now we will turn out attention to the second and the more important, the
model itself. We will start with a simple neural network with one hidden
layer and will motivate it using a problem of binary classification on a
simulated data set. We start by generating a simple dataset shown in
Figure below. The data is generated from a mixture of two distributions
(Gaussian and truncated Gaussian). The red points are the positive class
and the green points are the negative class. The goal is to find a model
boundary that discriminates the two classes.

\begin{center}\includegraphics[width=0.7\linewidth]{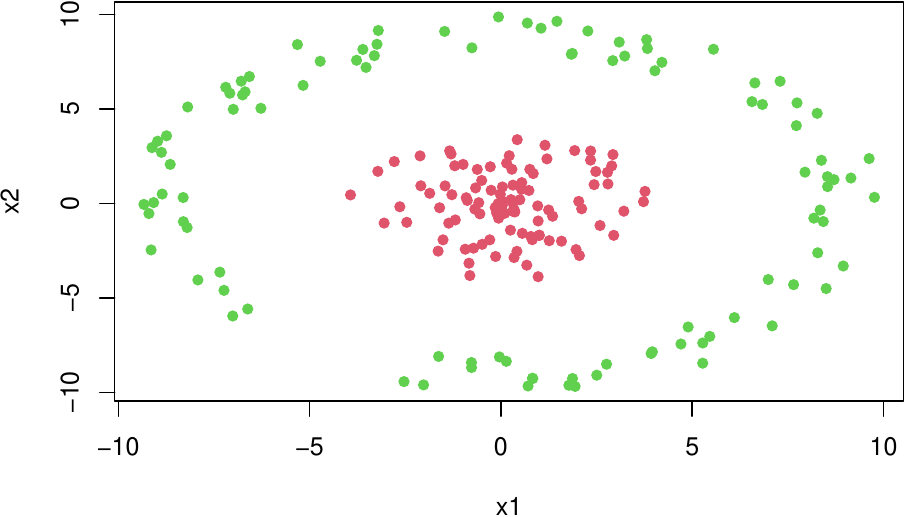} \end{center}

We can see that a logistic regression could not do it. It uses a single
line to separate observations of two classes.

\begin{Shaded}
\begin{Highlighting}[]
\CommentTok{\# Fit a logistic regression model}
\NormalTok{fit }\OtherTok{=} \FunctionTok{glm}\NormalTok{(label}\SpecialCharTok{\textasciitilde{}}\NormalTok{x1}\SpecialCharTok{+}\NormalTok{x2, }\AttributeTok{data=}\FunctionTok{as.data.frame}\NormalTok{(d), }\AttributeTok{family=}\FunctionTok{binomial}\NormalTok{(}\AttributeTok{link=}\StringTok{\textquotesingle{}logit\textquotesingle{}}\NormalTok{))}
\CommentTok{\# Plot the training dataset}
\FunctionTok{plot}\NormalTok{(d[,}\DecValTok{2}\NormalTok{],d[,}\DecValTok{3}\NormalTok{], }\AttributeTok{col=}\NormalTok{d[,}\DecValTok{1}\NormalTok{]}\SpecialCharTok{+}\DecValTok{2}\NormalTok{, }\AttributeTok{pch=}\DecValTok{16}\NormalTok{, }\AttributeTok{xlab=}\StringTok{"x1"}\NormalTok{, }\AttributeTok{ylab=}\StringTok{"x2"}\NormalTok{)}
\NormalTok{th }\OtherTok{=}\NormalTok{ fit}\SpecialCharTok{$}\NormalTok{coefficients}
\CommentTok{\# Plot the decision boundary}
\FunctionTok{abline}\NormalTok{(}\SpecialCharTok{{-}}\NormalTok{th[}\DecValTok{1}\NormalTok{]}\SpecialCharTok{/}\NormalTok{th[}\DecValTok{3}\NormalTok{], }\SpecialCharTok{{-}}\NormalTok{th[}\DecValTok{2}\NormalTok{]}\SpecialCharTok{/}\NormalTok{th[}\DecValTok{3}\NormalTok{], }\AttributeTok{col=}\DecValTok{2}\NormalTok{)}
\end{Highlighting}
\end{Shaded}

\begin{center}\includegraphics[width=0.7\linewidth]{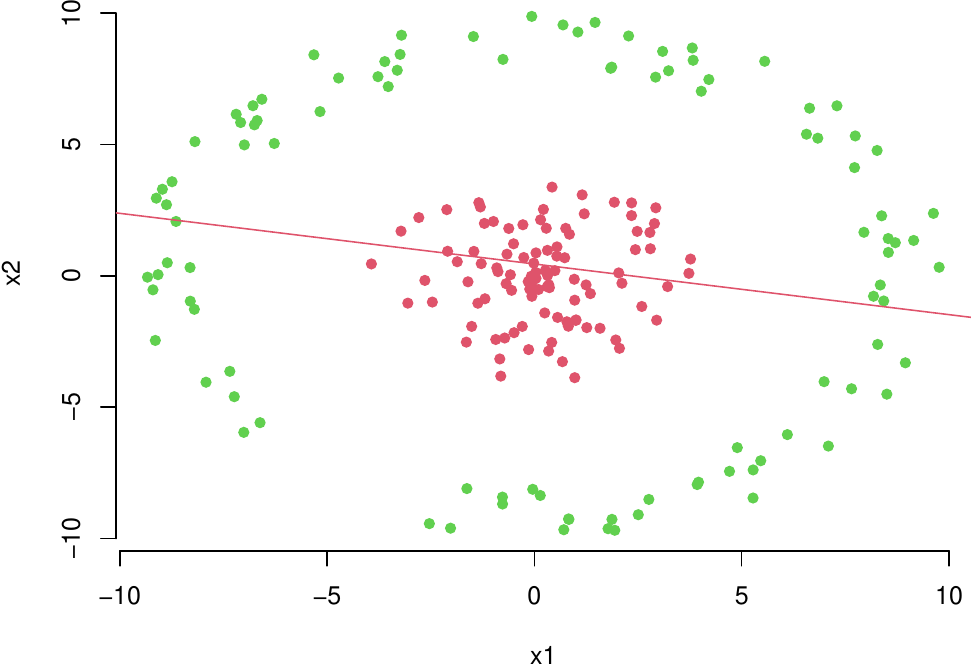} \end{center}

Indeed, the line found by the logistic regression is not able to
separate the two classes. We can see that the data is not linearly
separable. However, we can use multiple lines to separate the data.

\begin{Shaded}
\begin{Highlighting}[]
\FunctionTok{plot}\NormalTok{(x1}\SpecialCharTok{\textasciitilde{}}\NormalTok{x2, }\AttributeTok{data=}\NormalTok{d,}\AttributeTok{col=}\NormalTok{d[,}\DecValTok{1}\NormalTok{]}\SpecialCharTok{+}\DecValTok{2}\NormalTok{, }\AttributeTok{pch=}\DecValTok{16}\NormalTok{)}
\CommentTok{\# Plot lines that separate once class (red) from another (green)}
\FunctionTok{lines}\NormalTok{(x1, }\SpecialCharTok{{-}}\NormalTok{x1 }\SpecialCharTok{{-}} \DecValTok{6}\NormalTok{); }\FunctionTok{text}\NormalTok{(}\SpecialCharTok{{-}}\DecValTok{4}\NormalTok{,}\SpecialCharTok{{-}}\DecValTok{3}\NormalTok{,}\DecValTok{1}\NormalTok{)}
\FunctionTok{lines}\NormalTok{(x1, }\SpecialCharTok{{-}}\NormalTok{x1 }\SpecialCharTok{+} \DecValTok{6}\NormalTok{); }\FunctionTok{text}\NormalTok{(}\DecValTok{4}\NormalTok{,}\DecValTok{3}\NormalTok{,}\DecValTok{2}\NormalTok{)}
\FunctionTok{lines}\NormalTok{(x1,  x1 }\SpecialCharTok{{-}} \DecValTok{6}\NormalTok{); }\FunctionTok{text}\NormalTok{(}\DecValTok{4}\NormalTok{,}\SpecialCharTok{{-}}\DecValTok{3}\NormalTok{,}\DecValTok{3}\NormalTok{)}
\FunctionTok{lines}\NormalTok{(x1,  x1 }\SpecialCharTok{+} \DecValTok{6}\NormalTok{); }\FunctionTok{text}\NormalTok{(}\SpecialCharTok{{-}}\DecValTok{3}\NormalTok{,}\DecValTok{4}\NormalTok{,}\DecValTok{4}\NormalTok{)}
\end{Highlighting}
\end{Shaded}

\begin{center}\includegraphics[width=0.7\linewidth]{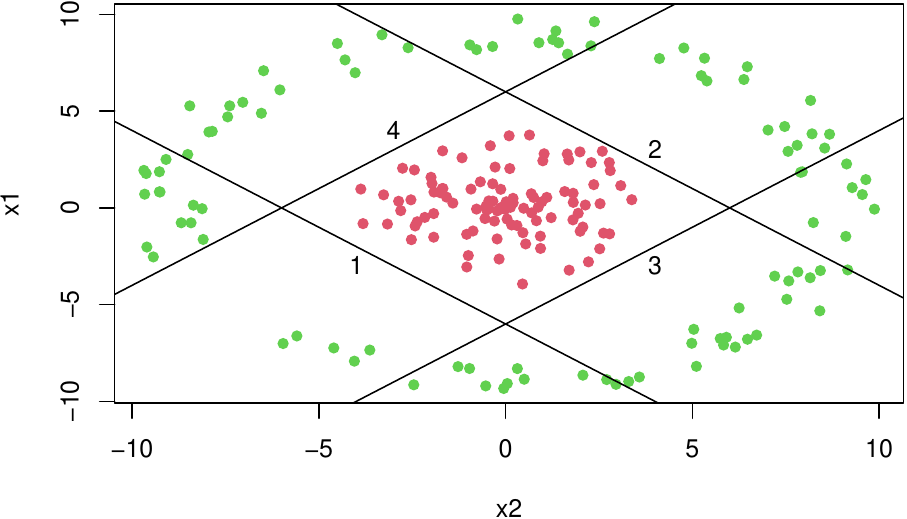} \end{center}

Now, we do the same thing as in simple logistic regression and apply
logistic function to each of those lines

\begin{Shaded}
\begin{Highlighting}[]
\CommentTok{\# Define sigmoid function}
\NormalTok{sigmoid  }\OtherTok{=} \ControlFlowTok{function}\NormalTok{(z)}
  \FunctionTok{return}\NormalTok{(}\FunctionTok{exp}\NormalTok{(z)}\SpecialCharTok{/}\NormalTok{(}\DecValTok{1}\SpecialCharTok{+}\FunctionTok{exp}\NormalTok{(z)))}

\CommentTok{\# Define hidden layer of our neural network}
\NormalTok{features }\OtherTok{=} \ControlFlowTok{function}\NormalTok{(x1,x2) \{}
\NormalTok{  z1 }\OtherTok{=}  \DecValTok{6} \SpecialCharTok{+}\NormalTok{ x1 }\SpecialCharTok{+}\NormalTok{ x2; a1 }\OtherTok{=} \FunctionTok{sigmoid}\NormalTok{(z1)}
\NormalTok{  z2 }\OtherTok{=}  \DecValTok{6} \SpecialCharTok{{-}}\NormalTok{ x1 }\SpecialCharTok{{-}}\NormalTok{ x2; a2 }\OtherTok{=} \FunctionTok{sigmoid}\NormalTok{(z2)}
\NormalTok{  z3 }\OtherTok{=}  \DecValTok{6} \SpecialCharTok{{-}}\NormalTok{ x1 }\SpecialCharTok{+}\NormalTok{ x2; a3 }\OtherTok{=} \FunctionTok{sigmoid}\NormalTok{(z3)}
\NormalTok{  z4 }\OtherTok{=}  \DecValTok{6} \SpecialCharTok{+}\NormalTok{ x1 }\SpecialCharTok{{-}}\NormalTok{ x2; a4 }\OtherTok{=} \FunctionTok{sigmoid}\NormalTok{(z4)}
  \FunctionTok{return}\NormalTok{(}\FunctionTok{c}\NormalTok{(a1,a2,a3,a4))}
\NormalTok{\}}
\end{Highlighting}
\end{Shaded}

Using the matrix notaitons, we have \[
z = \sigma(Wx + b), ~ W = \begin{bmatrix} 1 & 1 \\ -1 & -1 \\ -1 & 1 \\ 1 & -1 \end{bmatrix}, ~ b = \begin{bmatrix} 6 \\ 6 \\ 6 \\ 6 \end{bmatrix}, ~ \sigma(z) = \frac{1}{1+e^{-z}}
\]

The model shown above is the first layer of our neural network. It takes
a two-dimensional input \(x\) and produces a four-dimensional output
\(z\) which is a called a feature vector. The feature vector is then
passed to the output layer, which applies simple logistic regression to
the feature vector. \[
\hat{y} = \sigma(w^Tz + b), ~ w = \begin{bmatrix} 1 \\ 1 \\ 1 \\ 1 \end{bmatrix}, ~ b = -3.1, ~ \sigma(z) = \frac{1}{1+e^{-z}}
\]

The output of the output layer is the probability of the positive class.

\begin{Shaded}
\begin{Highlighting}[]
\CommentTok{\# Calculate prediction (classification) using our neural network}
\NormalTok{predict\_prob }\OtherTok{=} \ControlFlowTok{function}\NormalTok{(x)\{}
\NormalTok{  x1 }\OtherTok{=}\NormalTok{ x[}\DecValTok{1}\NormalTok{]; x2 }\OtherTok{=}\NormalTok{ x[}\DecValTok{2}\NormalTok{]}
\NormalTok{  z }\OtherTok{=} \FunctionTok{features}\NormalTok{(x1,x2)}
  \CommentTok{\# print(z)}
\NormalTok{  mu }\OtherTok{=} \FunctionTok{sum}\NormalTok{(z) }\SpecialCharTok{{-}} \FloatTok{3.1}
  \CommentTok{\# print(mu)}
  \FunctionTok{sigmoid}\NormalTok{(mu)}
\NormalTok{\}}
\end{Highlighting}
\end{Shaded}

We can use our model to do the predictions now

\begin{Shaded}
\begin{Highlighting}[]
\CommentTok{\# Predict the probability of the positive class for a given point}
\FunctionTok{predict\_prob}\NormalTok{(}\FunctionTok{c}\NormalTok{(}\DecValTok{0}\NormalTok{,}\DecValTok{0}\NormalTok{))}
\CommentTok{\#\textgreater{} [1] 0.71}
\FunctionTok{predict\_prob}\NormalTok{(}\FunctionTok{c}\NormalTok{(}\DecValTok{0}\NormalTok{,}\DecValTok{10}\NormalTok{))}
\CommentTok{\#\textgreater{} [1] 0.26}
\end{Highlighting}
\end{Shaded}

The model generates sensible predictions, let's plot the decision
boundary to see how well it separates the data.

\begin{Shaded}
\begin{Highlighting}[]
\NormalTok{x1 }\OtherTok{=} \FunctionTok{seq}\NormalTok{(}\SpecialCharTok{{-}}\DecValTok{11}\NormalTok{,}\DecValTok{11}\NormalTok{,}\AttributeTok{length.out =} \DecValTok{100}\NormalTok{)}
\NormalTok{x2 }\OtherTok{=} \FunctionTok{seq}\NormalTok{(}\SpecialCharTok{{-}}\DecValTok{11}\NormalTok{,}\DecValTok{11}\NormalTok{,}\AttributeTok{length.out =} \DecValTok{100}\NormalTok{)}
\NormalTok{gr }\OtherTok{=} \FunctionTok{as.matrix}\NormalTok{(}\FunctionTok{expand.grid}\NormalTok{(x1,x2));}
\CommentTok{\#\textgreater{} [1] 10000     2}
\NormalTok{yhat }\OtherTok{=} \FunctionTok{apply}\NormalTok{(gr,}\DecValTok{1}\NormalTok{,predict\_prob)}
\CommentTok{\#\textgreater{} [1] 10000}
\FunctionTok{image}\NormalTok{(x1,x2,}\FunctionTok{matrix}\NormalTok{(yhat,}\AttributeTok{ncol =} \DecValTok{100}\NormalTok{), }\AttributeTok{col =} \FunctionTok{heat.colors}\NormalTok{(}\DecValTok{20}\NormalTok{,}\FloatTok{0.7}\NormalTok{))}
\end{Highlighting}
\end{Shaded}

\begin{center}\includegraphics[width=0.7\linewidth]{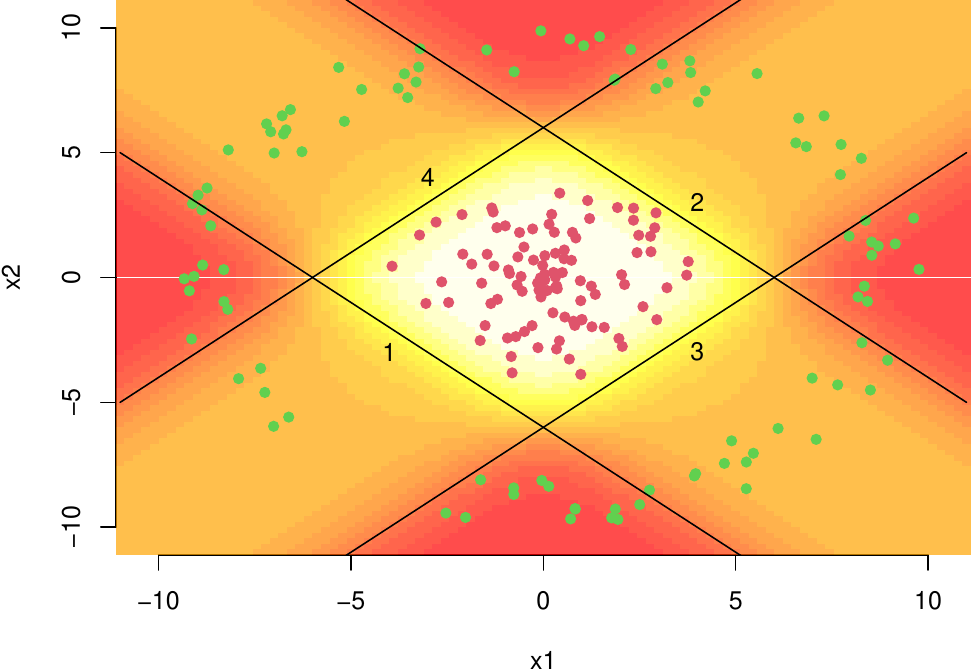} \end{center}

\hypertarget{automatic-differentiation-backpropagation}{%
\section{Automatic Differentiation
(Backpropagation)}\label{automatic-differentiation-backpropagation}}

To calculate the value of the gradient vector, at each step of the
optimization process, deep learning libraries require calculations of
derivatives. In general, there are three different ways to calculate
those derivatives. First, is numerical differentiation, when a gradient
is approximated by a finite difference \(f'(x) = (f(x+h)-f(x))/h\) and
requires two function evaluations. However, the numerical
differentiation is not backward stable (Griewank et al. 2012), meaning
that for a small perturbation in input value \(x\), the calculated
derivative is not the correct one. Second, is a symbolic differentiation
which has been used in symbolic computational frameworks such as
\texttt{Mathematica} or \texttt{Maple} for decades. Symbolic
differentiation uses a tree form representation of a function and
applies chain rule to the tree to calculate the symbolic derivative of a
given function. Figure @ref(fig:comp-graph) shows a tree representation
of of composition of affine and sigmoid functions (the first layer of
our neural network).

\begin{figure}

{\centering \includegraphics[width=1\linewidth]{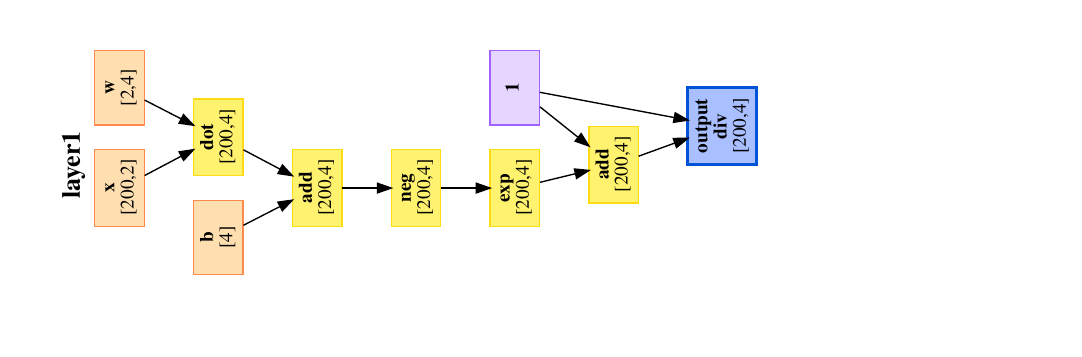} 

}

\caption{Computational graph of the first layer of our neural network}\label{fig:comp-graph}
\end{figure}

The advantage of symbolic calculations is that the analytical
representation of derivative is available for further analysis. For
example, when derivative calculation is in an intermediate step of the
analysis. Third way to calculate a derivative is to use automatic
differentiation (AD). Similar to symbolic differentiation, AD
recursively applies the chain rule and calculates the exact value of
derivative and thus avoids the problem of numerical instability. The
difference between AD and symbolic differentiation is that AD provides
the value of derivative evaluated at a specific point, rather than an
analytical representation of the derivative.

AD does not require analytical specification and can be applied to a
function defined by a sequence of algebraic manipulations, logical and
transient functions applied to input variables and specified in a
computer code. AD can differentiate complex functions which involve IF
statements and loops, and AD can be implemented using either forward or
backward mode. Consider an example of calculating a derivative of the
following function with respect to \texttt{x}.

\begin{Shaded}
\begin{Highlighting}[]
\NormalTok{sigmoid }\OtherTok{=} \ControlFlowTok{function}\NormalTok{(x,b,w)\{}
\NormalTok{  v1 }\OtherTok{=}\NormalTok{ w}\SpecialCharTok{*}\NormalTok{x;}
\NormalTok{  v2 }\OtherTok{=}\NormalTok{ v1 }\SpecialCharTok{+}\NormalTok{ b}
\NormalTok{  v3 }\OtherTok{=} \DecValTok{1}\SpecialCharTok{/}\NormalTok{(}\DecValTok{1}\SpecialCharTok{+}\FunctionTok{exp}\NormalTok{(}\SpecialCharTok{{-}}\NormalTok{v2))}
\NormalTok{\}}
\end{Highlighting}
\end{Shaded}

In the forward mode an auxiliary variable, called a dual number, will be
added to each line of the code to track the value of the derivative
associated with this line. In our example, if we set
\texttt{x=2,\ w=3,\ b=5}2, we get the calculations given in Table below.

\begin{longtable}[]{@{}
  >{\raggedright\arraybackslash}p{(\columnwidth - 2\tabcolsep) * \real{0.3511}}
  >{\raggedright\arraybackslash}p{(\columnwidth - 2\tabcolsep) * \real{0.6489}}@{}}
\toprule\noalign{}
\begin{minipage}[b]{\linewidth}\raggedright
Function calculations
\end{minipage} & \begin{minipage}[b]{\linewidth}\raggedright
Derivative calculations
\end{minipage} \\
\midrule\noalign{}
\endhead
\bottomrule\noalign{}
\endlastfoot
1. \texttt{v1\ =\ w*x\ =\ 6} & 1. \texttt{dv1\ =\ w\ =\ 3} (derivative
of \texttt{v1} with respect to \texttt{x}) \\
2. \texttt{v2\ =\ v1\ +\ b\ =\ 11} & 2. \texttt{dv2\ =\ dv1\ =\ 3}
(derivative of \texttt{v2} with respect to \texttt{x}) \\
3. \texttt{v3\ =\ 1/(1+exp(-v2))\ =\ 0.99} & 3.
\texttt{dv3\ =\ eps2*exp(-v2)/(1+exp(-v2))**2\ \ =\ 5e-05} \\
\end{longtable}

Variables \texttt{dv1,dv2,dv3} correspond to partial (local) derivatives
of each intermediate variables \texttt{v1,v2,v3} with respect to \(x\),
and are called dual variables. Tracking for dual variables can either be
implemented using source code modification tools that add new code for
calculating the dual numbers or via operator overloading.

The reverse AD also applies chain rule recursively but starts from the
outer function, as shown in Table below.

\begin{longtable}[]{@{}
  >{\raggedright\arraybackslash}p{(\columnwidth - 2\tabcolsep) * \real{0.3929}}
  >{\raggedright\arraybackslash}p{(\columnwidth - 2\tabcolsep) * \real{0.6071}}@{}}
\toprule\noalign{}
\begin{minipage}[b]{\linewidth}\raggedright
Function calculations
\end{minipage} & \begin{minipage}[b]{\linewidth}\raggedright
Derivative calculations
\end{minipage} \\
\midrule\noalign{}
\endhead
\bottomrule\noalign{}
\endlastfoot
1. \texttt{v1\ =\ w*x\ =\ 6} & 4.
\texttt{dv1dx\ =w;\ dv1\ =\ dv2*dv1dx\ =\ 3*1.3e-05=5e-05} \\
2. \texttt{v2\ =\ v1\ +\ b\ =\ 11} & 3.
\texttt{dv2dv1\ =1;\ dv2\ =\ dv3*dv2dv1\ =\ 1.3e-05} \\
3. \texttt{v3\ =\ 1/(1+exp(-v2))\ =\ 0.99} & 2.
\texttt{dv3dv2\ =\ exp(-v2)/(1+exp(-v2))**2;} \\
4. \texttt{v4\ =\ v3} & 1. \texttt{dv4=1} \\
\end{longtable}

For DL, derivatives are calculated by applying reverse AD algorithm to a
model which is defined as a superposition of functions. A model is
defined either using a general purpose language as it is done in
\texttt{PyTorch} or through a sequence of function calls defined by
framework libraries (e.g.~in \texttt{TensorFlow}). Forward AD algorithms
calculate the derivative with respect to a single input variable, but
reverse AD produces derivatives with respect to all intermediate
variables. For models with many parameters, it is much more
computationally feasible to perform the reverse AD.

In the context of neural networks, the reverse AD algorithms is called
back-propagation and was popularized in AI by Rumelhart et al. (1986).
According to Schmidhuber (2015) the first version of what we call today
back-propagation was published in 1970 in a master's thesis Linnainmaa
(1970) and was closely related to the work of Ostrovskii et al. (1971).
However, similar techniques rooted in Pontryagin's maximization
principle Boltyanskii et al. (1960) were discussed in the context of
multi-stage control problems Bryson (1961),bryson1969applied\}. Dreyfus
(1962) applies back-propagation to calculate the first order derivative
of a return function to numerically solve a variational problem. Later
Dreyfus (1973) used back-propagation to derive an efficient algorithm to
solve a minimization problem. The first neural network specific version
of back-propagation was proposed in Werbos (1974) and an efficient
back-propagation algorithm was discussed in Werbos (1982).

Modern deep learning frameworks fully automate the process of finding
derivatives using AD algorithms. For example, \texttt{PyTorch} relies on
\texttt{autograd} library which automatically finds gradient using
back-propagation algorithm. Here is a small code example using
\texttt{autograd} library in \texttt{jax}.

\begin{Shaded}
\begin{Highlighting}[]
\ImportTok{import}\NormalTok{ jax.numpy }\ImportTok{as}\NormalTok{ jnp}
\ImportTok{from}\NormalTok{ jax }\ImportTok{import}\NormalTok{ grad,jit}
\ImportTok{import}\NormalTok{ pandas }\ImportTok{as}\NormalTok{ pd}
\ImportTok{from}\NormalTok{ jax }\ImportTok{import}\NormalTok{ random}
\ImportTok{import}\NormalTok{ matplotlib.pyplot }\ImportTok{as}\NormalTok{ plt}

\KeywordTok{def}\NormalTok{ abline(slope, intercept):}
    \CommentTok{"""Plot a line from slope and intercept"""}
\NormalTok{    axes }\OperatorTok{=}\NormalTok{ plt.gca()}
\NormalTok{    x\_vals }\OperatorTok{=}\NormalTok{ jnp.array(axes.get\_xlim())}
\NormalTok{    ylim }\OperatorTok{=}\NormalTok{ axes.get\_xlim()}
\NormalTok{    y\_vals }\OperatorTok{=}\NormalTok{ intercept }\OperatorTok{+}\NormalTok{ slope }\OperatorTok{*}\NormalTok{ x\_vals}
\NormalTok{    plt.plot(x\_vals, y\_vals, }\StringTok{\textquotesingle{}{-}\textquotesingle{}}\NormalTok{)}\OperatorTok{;}\NormalTok{ plt.ylim(ylim)}

\NormalTok{d }\OperatorTok{=}\NormalTok{ pd.read\_csv(}\StringTok{\textquotesingle{}circle.csv\textquotesingle{}}\NormalTok{).values}
\NormalTok{x }\OperatorTok{=}\NormalTok{ d[:, }\DecValTok{1}\NormalTok{:}\DecValTok{3}\NormalTok{]}\OperatorTok{;}\NormalTok{ y }\OperatorTok{=}\NormalTok{ d[:, }\DecValTok{0}\NormalTok{]}
\KeywordTok{def}\NormalTok{ sigmoid(x):}
    \ControlFlowTok{return} \DecValTok{1} \OperatorTok{/}\NormalTok{ (}\DecValTok{1} \OperatorTok{+}\NormalTok{ jnp.exp(}\OperatorTok{{-}}\NormalTok{x))}
\KeywordTok{def}\NormalTok{ predict(x, w1,b1,w2,b2):}
\NormalTok{    z }\OperatorTok{=}\NormalTok{ sigmoid(jnp.dot(x, w1)}\OperatorTok{+}\NormalTok{b1)}
    \ControlFlowTok{return}\NormalTok{ sigmoid(jnp.dot(z, w2)}\OperatorTok{+}\NormalTok{b2)[:,}\DecValTok{0}\NormalTok{]}
\KeywordTok{def}\NormalTok{ nll(x, y, w1,b1,w2,b2):}
\NormalTok{    yhat }\OperatorTok{=}\NormalTok{ predict(x, w1,b1,w2,b2)}
    \ControlFlowTok{return} \OperatorTok{{-}}\NormalTok{jnp.}\BuiltInTok{sum}\NormalTok{(y }\OperatorTok{*}\NormalTok{ jnp.log(yhat) }\OperatorTok{+}\NormalTok{ (}\DecValTok{1} \OperatorTok{{-}}\NormalTok{ y) }\OperatorTok{*}\NormalTok{ jnp.log(}\DecValTok{1} \OperatorTok{{-}}\NormalTok{ yhat))}
\AttributeTok{@jit}
\KeywordTok{def}\NormalTok{ sgd\_step(x, y, w1,b1,w2,b2, lr):}
\NormalTok{    grads }\OperatorTok{=}\NormalTok{ grad(nll,argnums}\OperatorTok{=}\NormalTok{[}\DecValTok{2}\NormalTok{,}\DecValTok{3}\NormalTok{,}\DecValTok{4}\NormalTok{,}\DecValTok{5}\NormalTok{])(x, y, w1,b1,w2,b2)}
    \ControlFlowTok{return}\NormalTok{ w1 }\OperatorTok{{-}}\NormalTok{ lr }\OperatorTok{*}\NormalTok{ grads[}\DecValTok{0}\NormalTok{],b1 }\OperatorTok{{-}}\NormalTok{ lr }\OperatorTok{*}\NormalTok{ grads[}\DecValTok{1}\NormalTok{],w2 }\OperatorTok{{-}}\NormalTok{ lr }\OperatorTok{*}\NormalTok{ grads[}\DecValTok{2}\NormalTok{],b2 }\OperatorTok{{-}}\NormalTok{ lr }\OperatorTok{*}\NormalTok{ grads[}\DecValTok{3}\NormalTok{]}
\KeywordTok{def}\NormalTok{ accuracy(x, y, w1,b1,w2,b2):}
\NormalTok{    y\_pred }\OperatorTok{=}\NormalTok{ predict(x, w1,b1,w2,b2)}
    \ControlFlowTok{return}\NormalTok{ jnp.mean((y\_pred }\OperatorTok{\textgreater{}} \FloatTok{0.5}\NormalTok{) }\OperatorTok{==}\NormalTok{ y)}
\NormalTok{k }\OperatorTok{=}\NormalTok{ random.PRNGKey(}\DecValTok{0}\NormalTok{)}
\NormalTok{w1 }\OperatorTok{=} \FloatTok{0.1}\OperatorTok{*}\NormalTok{random.normal(k,(}\DecValTok{2}\NormalTok{,}\DecValTok{4}\NormalTok{))}
\NormalTok{b1 }\OperatorTok{=} \FloatTok{0.01}\OperatorTok{*}\NormalTok{random.normal(k,(}\DecValTok{4}\NormalTok{,))}
\NormalTok{w2 }\OperatorTok{=} \FloatTok{0.1}\OperatorTok{*}\NormalTok{random.normal(k,(}\DecValTok{4}\NormalTok{,}\DecValTok{1}\NormalTok{))}
\NormalTok{b2 }\OperatorTok{=} \FloatTok{0.01}\OperatorTok{*}\NormalTok{random.normal(k,(}\DecValTok{1}\NormalTok{,))}

\ControlFlowTok{for}\NormalTok{ i }\KeywordTok{in} \BuiltInTok{range}\NormalTok{(}\DecValTok{1000}\NormalTok{):}
\NormalTok{    w1,b1,w2,b2 }\OperatorTok{=}\NormalTok{ sgd\_step(x,y,w1,b1,w2,b2,}\FloatTok{0.003}\NormalTok{)}
\BuiltInTok{print}\NormalTok{(accuracy(x,y,w1,b1,w2,b2))}
\CommentTok{\#\textgreater{} 1.0}

\NormalTok{fig, ax }\OperatorTok{=}\NormalTok{ plt.subplots()}
\NormalTok{ax.scatter(x[:,}\DecValTok{0}\NormalTok{], x[:,}\DecValTok{1}\NormalTok{], c}\OperatorTok{=}\NormalTok{[}\StringTok{\textquotesingle{}r\textquotesingle{}} \ControlFlowTok{if}\NormalTok{ x}\OperatorTok{==}\DecValTok{1} \ControlFlowTok{else} \StringTok{\textquotesingle{}g\textquotesingle{}} \ControlFlowTok{for}\NormalTok{ x }\KeywordTok{in}\NormalTok{ y],s}\OperatorTok{=}\DecValTok{7}\NormalTok{)}\OperatorTok{;}\NormalTok{ plt.xlabel(}\StringTok{"x1"}\NormalTok{)}\OperatorTok{;}\NormalTok{ plt.ylabel(}\StringTok{"x2"}\NormalTok{)}\OperatorTok{;}\NormalTok{ plt.xlim(}\OperatorTok{{-}}\DecValTok{10}\NormalTok{,}\DecValTok{10}\NormalTok{)}
\CommentTok{\#\textgreater{} \textless{}matplotlib.collections.PathCollection object at 0x1654b2f10\textgreater{}}
\CommentTok{\#\textgreater{} Text(0.5, 0, \textquotesingle{}x1\textquotesingle{})}
\CommentTok{\#\textgreater{} Text(0, 0.5, \textquotesingle{}x2\textquotesingle{})}
\CommentTok{\#\textgreater{} ({-}10.0, 10.0)}
\NormalTok{ax.spines[}\StringTok{\textquotesingle{}top\textquotesingle{}}\NormalTok{].set\_visible(}\VariableTok{False}\NormalTok{)}
\CommentTok{\# plt.scatter((x[:,1]*w1[1,0] {-} b1[0])/w1[0,0], x[:,1])}
\NormalTok{abline(w1[}\DecValTok{1}\NormalTok{,}\DecValTok{0}\NormalTok{]}\OperatorTok{/}\NormalTok{w1[}\DecValTok{0}\NormalTok{,}\DecValTok{0}\NormalTok{],b1[}\DecValTok{0}\NormalTok{]}\OperatorTok{/}\NormalTok{w1[}\DecValTok{0}\NormalTok{,}\DecValTok{0}\NormalTok{])}
\NormalTok{abline(w1[}\DecValTok{1}\NormalTok{,}\DecValTok{1}\NormalTok{]}\OperatorTok{/}\NormalTok{w1[}\DecValTok{0}\NormalTok{,}\DecValTok{1}\NormalTok{],b1[}\DecValTok{1}\NormalTok{]}\OperatorTok{/}\NormalTok{w1[}\DecValTok{0}\NormalTok{,}\DecValTok{1}\NormalTok{])}
\NormalTok{abline(w1[}\DecValTok{1}\NormalTok{,}\DecValTok{2}\NormalTok{]}\OperatorTok{/}\NormalTok{w1[}\DecValTok{0}\NormalTok{,}\DecValTok{2}\NormalTok{],b1[}\DecValTok{2}\NormalTok{]}\OperatorTok{/}\NormalTok{w1[}\DecValTok{0}\NormalTok{,}\DecValTok{2}\NormalTok{])}
\NormalTok{abline(w1[}\DecValTok{1}\NormalTok{,}\DecValTok{3}\NormalTok{]}\OperatorTok{/}\NormalTok{w1[}\DecValTok{0}\NormalTok{,}\DecValTok{3}\NormalTok{],b1[}\DecValTok{3}\NormalTok{]}\OperatorTok{/}\NormalTok{w1[}\DecValTok{0}\NormalTok{,}\DecValTok{3}\NormalTok{])}
\NormalTok{plt.show()}
\end{Highlighting}
\end{Shaded}

\begin{center}\includegraphics[width=0.7\linewidth]{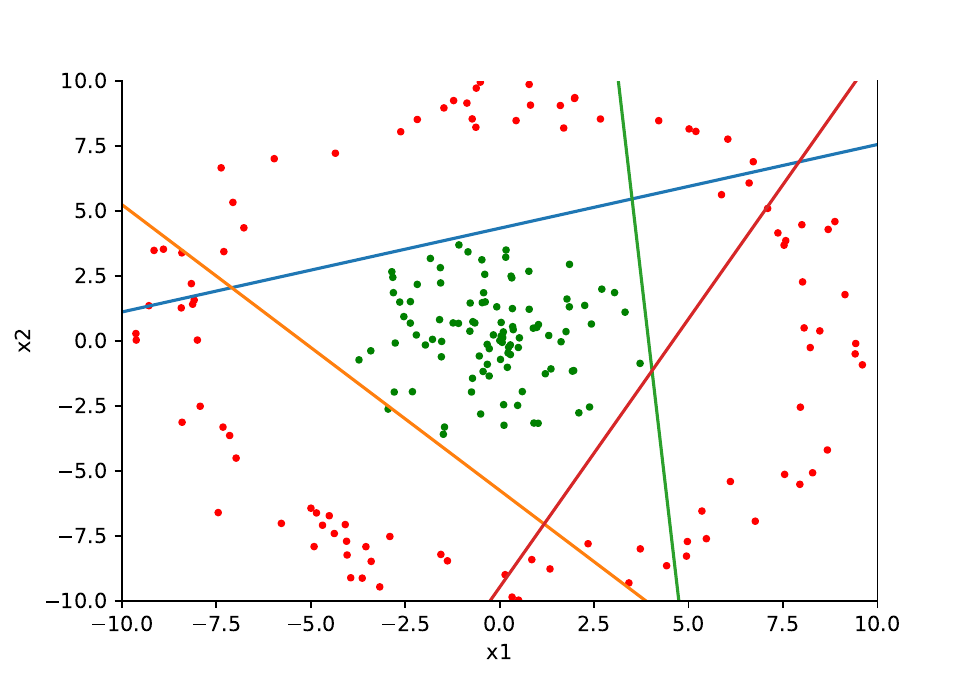} \end{center}

\hypertarget{discussion}{%
\section{Discussion}\label{discussion}}

The goal of our paper is to provide an overview of of DL for
statisticians. To do this, we have discussed the model estimation
procedure and demonstrated that DL is an extension of a generalized
linear model. One goal of statistics is to build predictive models along
with uncertainty and to develop an understanding of the data generating
mechanism. Data models are well studied in statistical literature, but
often do not provide enough flexibility to learn the input-output
relations. Closed box predictive rules, such as trees and neural
networks, are more flexible learners. However, in high-dimensional
problems, finding good models is challenging, and this is where deep
learning methods shine. We can think of deterministic DL model as a
transformation of high dimensional input and outputs. Hidden features
lie on the transformed space, and are empirically learned as opposed to
theoretically specified.

Although DL models have been almost exclusively used for problems of
image analysis and natural language processing, more traditional data
sets, which arise in finance, science and engineering, such as spatial
and temporal data can be efficiently analyzed using deep learning. Thus,
DL provides an alternative for applications where traditional
statistical techniques apply. There are a number of areas of future
research for Statisticians. In particular, uncertainty quantification
and model selection, such as architecture design, as well as algorithmic
improvements and Bayesian deep learning. We hope this review will make
DL models accessible for statisticians.

\hypertarget{references}{%
\section*{References}\label{references}}
\addcontentsline{toc}{section}{References}

\hypertarget{refs}{}
\begin{CSLReferences}{1}{0}
\leavevmode\vadjust pre{\hypertarget{ref-behnia2021deep}{}}%
Behnia F, Karbowski D, Sokolov V (2021) Deep generative models for
vehicle speed trajectories. arXiv preprint arXiv:211208361

\leavevmode\vadjust pre{\hypertarget{ref-bhadra2021merging}{}}%
Bhadra A, Datta J, Polson N, et al (2021) Merging two cultures: Deep and
statistical learning. arXiv preprint arXiv:211011561

\leavevmode\vadjust pre{\hypertarget{ref-BolGamPon60}{}}%
Boltyanskii VG, Gamkrelidze RV, Pontryagin (1960) Theory of optimal
processes i: Maximum principle. News of Akad Nauk SSSR Mathematics
Series 24:3--42

\leavevmode\vadjust pre{\hypertarget{ref-bottou2018optimization}{}}%
Bottou L, Curtis FE, Nocedal J (2018) Optimization methods for
large-scale machine learning. SIAM review 60:223--311

\leavevmode\vadjust pre{\hypertarget{ref-bryson1961gradient}{}}%
Bryson AE (1961) A gradient method for optimizing multi-stage allocation
processes. In: Proc. Harvard univ. Symposium on digital computers and
their applications

\leavevmode\vadjust pre{\hypertarget{ref-dixon2019deep}{}}%
Dixon MF, Polson NG, Sokolov VO (2019) Deep learning for spatio-temporal
modeling: Dynamic traffic flows and high frequency trading. Applied
Stochastic Models in Business and Industry 35:788--807

\leavevmode\vadjust pre{\hypertarget{ref-dreyfus1973computational}{}}%
Dreyfus S (1973) The computational solution of optimal control problems
with time lag. IEEE Transactions on Automatic Control 18:383--385

\leavevmode\vadjust pre{\hypertarget{ref-dreyfus1962numerical}{}}%
Dreyfus S (1962) The numerical solution of variational problems. Journal
of Mathematical Analysis and Applications 5:30--45

\leavevmode\vadjust pre{\hypertarget{ref-griewank2012numerical}{}}%
Griewank A, Kulshreshtha K, Walther A (2012) On the numerical stability
of algorithmic differentiation. Computing 94:125--149

\leavevmode\vadjust pre{\hypertarget{ref-hardt2016train}{}}%
Hardt M, Recht B, Singer Y (2016) Train faster, generalize better:
Stability of stochastic gradient descent. In: International conference
on machine learning. PMLR, pp 1225--1234

\leavevmode\vadjust pre{\hypertarget{ref-heaton2017deep}{}}%
Heaton J, Polson N, Witte JH (2017) Deep learning for finance: Deep
portfolios. Applied Stochastic Models in Business and Industry 33:3--12

\leavevmode\vadjust pre{\hypertarget{ref-keskar2016large}{}}%
Keskar NS, Mudigere D, Nocedal J, et al (2016) On large-batch training
for deep learning: Generalization gap and sharp minima. arXiv preprint
arXiv:160904836

\leavevmode\vadjust pre{\hypertarget{ref-lecun2002efficient}{}}%
LeCun Y, Bottou L, Orr GB, Müller K-R (2002) Efficient backprop. In:
Neural networks: Tricks of the trade. Springer, pp 9--50

\leavevmode\vadjust pre{\hypertarget{ref-linnainmaa1970representation}{}}%
Linnainmaa S (1970) The representation of the cumulative rounding error
of an algorithm as a taylor expansion of the local rounding errors.
Master's Thesis (in Finnish), Univ Helsinki 6--7

\leavevmode\vadjust pre{\hypertarget{ref-nareklishvili2022deep}{}}%
Nareklishvili M, Polson N, Sokolov V (2022a) Deep partial least squares
for iv regression. arXiv preprint arXiv:220702612

\leavevmode\vadjust pre{\hypertarget{ref-nareklishvili2022feature}{}}%
Nareklishvili M, Polson N, Sokolov V (2022b) Feature selection for
personalized policy analysis. arXiv preprint arXiv:230100251

\leavevmode\vadjust pre{\hypertarget{ref-nareklishvili2023generative}{}}%
Nareklishvili M, Polson N, Sokolov V (2023) Generative causal inference.
arXiv preprint arXiv:230616096

\leavevmode\vadjust pre{\hypertarget{ref-ostrovskii1971uber}{}}%
Ostrovskii G, Volin YM, Borisov W (1971) Uber die berechnung von
ableitungen. Wissenschaftliche Zeitschrift der Technischen Hochschule f
ur Chemie, Leuna-Merseburg 13:382--384

\leavevmode\vadjust pre{\hypertarget{ref-polson2017deep}{}}%
Polson NG, Sokolov V (2017) Deep learning: A bayesian perspective

\leavevmode\vadjust pre{\hypertarget{ref-polson2023generative}{}}%
Polson NG, Sokolov V (2023) Generative AI for bayesian computation.
arXiv preprint arXiv:230514972

\leavevmode\vadjust pre{\hypertarget{ref-polson2020deep}{}}%
Polson N, Sokolov V (2020) Deep learning: Computational aspects. Wiley
Interdisciplinary Reviews: Computational Statistics 12:e1500

\leavevmode\vadjust pre{\hypertarget{ref-polson2021deep}{}}%
Polson N, Sokolov V, Xu J (2021) Deep learning partial least squares.
arXiv preprint arXiv:210614085

\leavevmode\vadjust pre{\hypertarget{ref-robbins1951stochastic}{}}%
Robbins H, Monro S (1951) {A Stochastic Approximation Method}. The
Annals of Mathematical Statistics 22:400--407.
\url{https://doi.org/10.1214/aoms/1177729586}

\leavevmode\vadjust pre{\hypertarget{ref-rumelhart1986learning}{}}%
Rumelhart DE, Hinton GE, Williams RJ (1986) Learning representations by
back-propagating errors. nature 323:533

\leavevmode\vadjust pre{\hypertarget{ref-schmidhuber2015deep}{}}%
Schmidhuber J (2015) Deep learning in neural networks: An overview.
Neural networks 61:85--117

\leavevmode\vadjust pre{\hypertarget{ref-sokolov2017discussion}{}}%
Sokolov V (2017) Discussion of 'deep learning for finance: Deep
portfolios'. Applied Stochastic Models in Business and Industry
33:16--18

\leavevmode\vadjust pre{\hypertarget{ref-wang2022data}{}}%
Wang Y, Polson N, Sokolov VO (2022) Data augmentation for bayesian deep
learning. Bayesian Analysis 1:1--29

\leavevmode\vadjust pre{\hypertarget{ref-werbos1974beyond}{}}%
Werbos P (1974) Beyond regression:" new tools for prediction and
analysis in the behavioral sciences. Ph D dissertation, Harvard
University

\leavevmode\vadjust pre{\hypertarget{ref-werbos1982applications}{}}%
Werbos PJ (1982) Applications of advances in nonlinear sensitivity
analysis. In: System modeling and optimization. Springer, pp 762--770

\end{CSLReferences}

\end{document}